\documentclass[conference]{IEEEtran}
\IEEEoverridecommandlockouts

\usepackage{cite}
\usepackage{amsmath,amssymb,amsfonts}
\usepackage{booktabs,makecell,tabularx}
\usepackage{algorithmic}
\usepackage{graphicx}
\usepackage{subcaption}
\usepackage{textcomp}
\usepackage{xcolor}

\def\BibTeX{{\rm B\kern-.05em{\sc i\kern-.025em b}\kern-.08em
    T\kern-.1667em\lower.7ex\hbox{E}\kern-.125emX}}
\begin{document}

\title{Efficient Multi-Object Tracking on Edge Devices via Reconstruction-Based Channel Pruning\\

\thanks{Supported by the Federal Ministry for Economic Affairs and Climate Action (BMWK) on the basis of a decision by the German Bundestag.}
}

\author{\IEEEauthorblockN{Jan Müller, Adrian Pigors}
\IEEEauthorblockA{\textit{Faculty IV, Department of Computer Science} \\
\textit{Hochschule Hannover -- University of Applied Sciences and Arts}\\
Hanover, Germany \\
\{jan.mueller, adrian.pigors\}@hs-hannover.de}
}

\maketitle

\begin{abstract}
The advancement of multi-object tracking (MOT) technologies presents the dual challenge of maintaining high performance while addressing critical security and privacy concerns. In applications such as pedestrian tracking, where sensitive personal data is involved, the potential for privacy violations and data misuse becomes a significant issue if data is transmitted to external servers. To mitigate these risks, processing data directly on an edge device—such as a smart camera—has emerged as a viable solution \cite{yrjanainen2020privacy,sun2020collaborative,park2023multi}. Edge computing ensures that sensitive information remains local, thereby aligning with stringent privacy principles and significantly reducing network latency.

However, the implementation of MOT on edge devices is not without its challenges. Edge devices typically possess limited computational resources, necessitating the development of highly optimized algorithms capable of delivering real-time performance under these constraints. The disparity between the computational requirements of state-of-the-art MOT algorithms and the capabilities of edge devices emphasizes a significant obstacle.

To address these challenges, we propose a neural network pruning method specifically tailored to compress complex networks, such as those used in modern MOT systems. This approach optimizes MOT performance by ensuring high accuracy and efficiency within the constraints of limited edge devices, such as NVIDIA's Jetson Orin Nano. 
By applying our pruning method, we achieve model size reductions of up to 70\% while maintaining a high level of accuracy and further improving performance on the Jetson Orin Nano, demonstrating the effectiveness of our approach for edge computing applications.

\end{abstract}

\section{Introduction}
Multi-object tracking is a challenging task that involves detecting multiple objects across a sequence of images while preserving their identities over time. The difficulty stems from the need to manage variations in object appearances and diverse motion patterns. For instance, tracking multiple pedestrians in a densely populated scene necessitates distinguishing between individuals with similar appearances, re-identifying them after occlusions, and accurately handling different motion dynamics such as varying walking speeds and directions. To address these challenges, modern MOT systems extensively utilize deep neural networks, with a particular emphasis on convolutional neural networks (CNNs) \cite{agrawal2024systematic}. CNNs are highly effective in learning and recognizing complex visual patterns, which are essential for accurate identity embeddings \cite{he2023fastreid} and object detection \cite{zhiqiang2017review}.

Nevertheless, CNN-based models frequently encounter difficulties in achieving real-time performance on off-the-shelf hardware, and even more so when deployed on edge devices \cite{vestias2019survey}. This represents a notable problem, as edge computing addresses many of the issues associated with contemporary MOT systems. By performing data processing locally, edge computing mitigates network latency, which is crucial for real-time applications such as autonomous driving, where delays can have critical safety implications \cite{liu2019edge}. In smart city applications, where pedestrian tracking is facilitated by smart cameras, edge computing enhances data privacy by reducing the need for extensive data transmission and keeping sensitive information processed locally \cite{yrjanainen2020privacy,sun2020collaborative,park2023multi}.

To address these efficiency challenges, researchers have employed various strategies, including developing specialized model architectures \cite{wang2019towards} and integrating more efficient object detectors into existing frameworks \cite{meimetis2023real}. However, these approaches often involve substantial modifications to the model architecture or integration framework.

In contrast, our research aims at compressing the network to enhance the efficiency of existing models without necessitating architectural overhauls. We focus on models based on the Joint Detection and Embedding (JDE) framework \cite{wang2019towards}, such as FairMOT \cite{zhang2021fairmot}, known for its balance between accuracy and efficiency. To improve efficiency, we apply structured channel pruning—a compressing technique that reduces memory footprint and computational complexity by removing entire channels from the model's weights. Structured channel pruning stands out among compression techniques due to its ability to deliver universal speedup without the need for specialized hardware or software frameworks \cite{he2023structured, cheng2023survey}.

\begin{figure}[t]
    \centering
    \includegraphics[width=\linewidth]{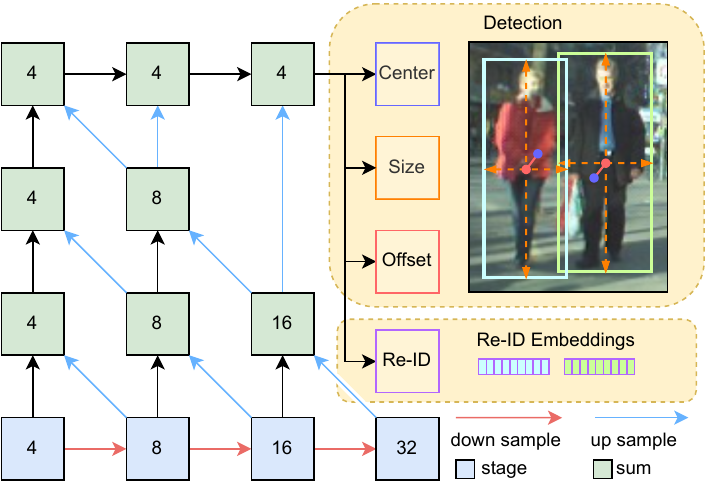}
    \caption{Overview of the FairMOT architecture. The model integrates object detection and re-identification (Re-ID) using CenterNet \cite{zhou2019objects} with the DLA-34  \cite{yu2018deep} backbone. It produces outputs for object center and class, object size, offset (correcting for quantization at a downsampled resolution of $ \frac{H}{4} \times \frac{W}{4} $, where $ H $ and $ W $ are the input image height and width), and Re-ID embedding. The numbers in the boxes refer to the downsampling factors relative to the original image resolution. (Based on \cite{yu2018deep, zhou2019objects, zhang2021fairmot}.)}
    \label{fig:fairmot_arch}
\end{figure}

However, implementing structured channel pruning presents significant challenges due to the interdependencies between different layers of the network \cite{fang2023depgraph}. For instance, pruning the output channels of a convolutional layer necessitates corresponding adjustments to the input channels of subsequent layers. This issue becomes particularly complex in modern models, such as those featured by JDE, which exhibit intricate and tightly coupled internal structures. FairMOT, as illustrated in Fig. \ref{fig:fairmot_arch}, exemplifies these complexities with its intricate architecture.

Many channel pruning methods, such as ThiNet \cite{luo2017thinet}—which employs a reconstruction-based approach—have tackled these issues through extensive, case-by-case analyses of various coupling scenarios. This approach often requires complicated, model-specific adjustments, making it both labor-intensive and inefficient. To mitigate these difficulties, we employ a Dependency Graph (DepGraph) \cite{fang2023depgraph} to prune groups of layers, managing the associated dependencies.

In this work, we introduce an innovative channel pruning technique that utilizes DepGraph for optimizing complex MOT networks on edge devices such as the Jetson Orin Nano. Specifically, we achieve:
\begin{enumerate}
    \item Development of a global and iterative reconstruction-based pruning pipeline. This pipeline can be applied to complex JDE-based networks, enabling the simultaneous pruning of both detection and re-identification components.
    \item Introduction of the gated groups concept, which enables the application of reconstruction-based pruning to groups of layers. This process also results in a more efficient pruning process by reducing the number of inference steps required for individual layers within a group. To our knowledge, this is the first application of reconstruction-based pruning criteria leveraging grouped layers.
    \item Our approach reduces the model’s parameters by 70\%, resulting in enhanced performance on the Jetson Orin Nano with minimal impact on accuracy. This highlights the practical efficiency and effectiveness of our pruning strategy on resource-constrained edge devices.
\end{enumerate}

\section{Related Work}
State-of-the-art multi-object tracking methods typically follow the tracking-by-detection paradigm \cite{zhang2021fairmot}. In this approach, objects are first detected in each frame, generating bounding boxes. To track these objects across frames, various association criteria are applied \cite{zhang2021fairmot}. For instance, location-based criteria might use a metric to assess the spatial overlap between bounding boxes. Motion-based criteria often rely on position estimates provided by the Kalman Filter \cite{welch1995introduction}. The criteria then involve calculating distances or overlaps between detections and estimates. Feature-based criteria might utilize re-identification embeddings to assess similarity between objects using measures like cosine similarity, ensuring consistent object identities across frames.

Recent research has focused not only on enhancing the accuracy of these tracking-by-detection methods, but also on improving their efficiency. Innovations in object detection have introduced models that are compact by design, such as MobileNet and Tiny-YOLO \cite{tsai2022mobilenet, wu2021sort, meimetis2023real}, which offer rapid inference times while maintaining robust performance. In addition to these inherently compact models, channel pruning techniques have been employed to enhance efficiency by selectively removing less important channels from pre-trained models \cite{mao2019efficient, jung2020real}. These advancements are complemented by improvements in the tracking pipeline itself. For instance, parallel Kalman Filters enable concurrent computations to accelerate tracking \cite{liu2024fasttrack}, while knowledge distillation techniques streamline Re-ID by creating more efficient models \cite{he2023fastreid}. Furthermore, frameworks like JDE integrate detection and embedding tasks into a unified model, optimizing both speed and accuracy by eliminating the need for separate processing stages. 

In our research, we focus on pruning JDE-based models like FairMOT, enabling the simultaneous pruning of both the re-identification network and the object detector.

\section{Pruning Methodology}
In this section, we describe our proposed reconstruction-based pruning method. We begin with the definition of reconstruction-based pruning, followed by our concept and utilization of gated groups. Finally, we showcase details of the iterative and global aspects of the pipeline.
\subsection{Reconstruction-based Pruning}
Reconstruction-based pruning involves removing specific input channels from a layer, while aiming to keep the output as close as possible to the original.

Consider the 2D output feature map produced by filter $f$ across the set of all input channels $C$ in a convolutional layer. We can approximate the output $O_f^C$ of the filter using a subset $C' \subseteq C$, i.\,e.
\begin{equation}
O_f^{C} \approx O_f^{C'}.
\end{equation}

To formalize this approximation for an entire layer, encompassing all filters, we can frame it as an optimization problem by minimizing the reconstruction error:

\begin{equation}
      \min_{C' \subseteq C} \sum_{f,i,j} \left ( O_f^{C}(i,j) - O_f^{C'}(i,j) \right )^2
     \label{eq:optimization_prob}
\end{equation}
 with $(i,j)$ denoting the spatial positions within the feature maps.

Since equation \eqref{eq:optimization_prob} is infeasible to compute, we adopt a greedy approach similar to ThiNet \cite{luo2017thinet}, where channels are selected by iteratively evaluating the impact of pruning specific input channels on the layer’s output. This method allows us to preserve the integrity of the feature representations within the layer, ensuring that critical information is maintained.
\subsection{Gated Groups}
Pruning targets are often grouped due to their interdependencies, necessitating a reconsideration of layer-wise reconstruction criteria. A natural approach is to aggregate the reconstruction error across all layers within a group. However, this approach has two significant drawbacks: first, it requires forward passes through each layer, which is computationally expensive; and second, it must address edge cases such as intradependencies in parameterized layers. For instance, when pruning a layer's input, one must also prune its output, as seen in depthwise convolutions, where a single convolutional filter is applied for each input channel \cite{chollet2017xception}.

\begin{figure}[t]
    \centering
    \begin{subfigure}[b]{0.455\linewidth} 
        \centering
        \includegraphics[width=\linewidth]{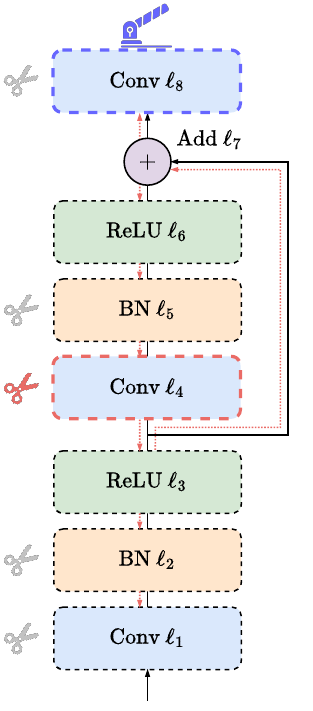}
        \caption{Gated Group in Computational Graph}
        \label{fig:sub1}
    \end{subfigure}
    \hfill
    \begin{subfigure}[b]{0.52\linewidth} 
        \centering
        \includegraphics[width=\linewidth]{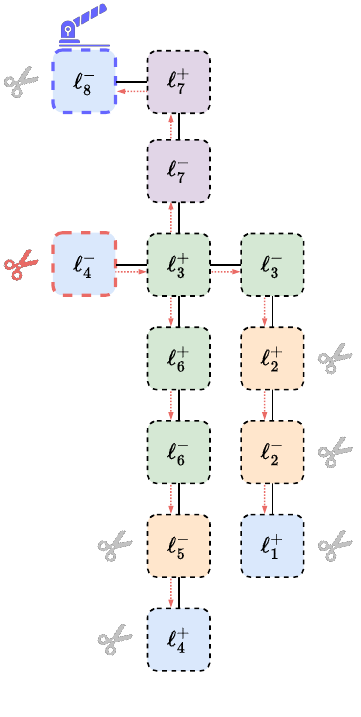}
        \caption{Gated Group in Dependency Graph}
        \label{fig:sub2}
    \end{subfigure}
    \caption{Illustration of a gated group with the convolutional layer \( \ell_4^- \) as the pruning target, where \( \ell_*^- \) denotes input channels and \( \ell_*^+ \) output channels for layer \( \ell_* \). In this example, \( \ell_8 \) is the only element within the gate set. The dashed red arrows highlight the propagation of dependencies originating from target \( \ell_4^- \). (Adapted from \cite[p. 4]{fang2023depgraph}.)}
    \label{fig:gated}
\end{figure}

To mitigate these issues, we propose a straightforward yet effective method. Instead of aggregating across all layers within a group, we focus on layers that act as gates, representing the endpoints of information flow within the group's computational graph (Fig. \ref{fig:gated}). By aggregating the reconstruction error over these gate layers, such as computing the mean reconstruction error, we effectively measure the impact of pruning the entire group.  In other words, we aggregate the reconstruction error over the gate set, which comprises the subset of all layers within a group that have no descendants in the computational graph within the given group. To further streamline this process, we automated the identification of gate layers by incorporating the computational graph of the network within DepGraph.

This approach not only simplifies the importance calculation by reducing the number of layers to consider but also eliminates the need to handle edge cases, as the output channels of gate layers cannot be pruning targets within their group.
\subsection{Iterative Pruning}
In our approach, we employ a global iterative pruning pipeline to systematically reduce the number of parameters in the model. In this context, global pruning involves evaluating all groups within the network collectively at each pruning step, rather than assessing them independently. This method ensures a more holistic reduction in parameters by considering the entire network as a whole.

The process follows a pre-defined pruning ratio, with a constant number of steps where, at each step, the number of parameters is linearly reduced. Each pruning step is followed by a training phase, managing the trade-off between model accuracy and sparsity. 

\section{Experiments and Results}

\subsection{Implementation Details}
We evaluate our methods using the MOT20 \cite{dendorfer2020mot20} dataset. To ensure consistent comparisons, we train our own FairMOT baseline model on top of the original pretrained FairMOT, which was initially trained on the CrowdHuman \cite{shao2018crowdhuman} and MIX \cite{wang2019towards} datasets.

Our training process involves an 80-20 split of the MOT20 and MOT17 \cite{milan2016mot16} datasets, stratified by re-identification labels. The 20 percent split is reserved for performance monitoring and early stopping to prevent overfitting. We apply data augmentation techniques, including random rotation, scaling, cropping, flipping, and color jittering, to enhance the diversity of the training data. The baseline model is trained with an input resolution of $1088 \times 608$, a batch size of 32, and a learning rate of $10^{-4}$, with training concluding at epoch 22.

During iterative pruning, we remove one percent of the model’s parameters per step and retrain for one epoch only if the validation loss increases relative to the previous step.

\subsection{Metrics}
We evaluate tracking performance using the CLEAR metrics \cite{bernardin2008evaluating}, including MOTA, FP, FN, and IDs, along with IDF1 \cite{ristani2016performance} and HOTA \cite{luiten2021hota}. MOTA emphasizes detection accuracy, driven by FP, FN, and IDs, while IDF1 focuses on identity preservation, highlighting association performance. HOTA explicitly balances
the effect of performing accurate detection, association and
localization into one metric.

\subsection{Benchmark Evaluation}
Both our baseline and pruned models are evaluated on the MOT challenge server under the private detector protocol. As shown in Table \ref{table:original_ours_baseline}, our baseline model outperforms the original FairMOT on several metrics. Moreover, we successfully pruned 70\% of the model’s parameters while maintaining a high level of accuracy. 

\begin{table}[!ht]
    \caption{Comparison of tracking performance between the original FairMOT and our pretrained and pruned baseline on the MOT20 test set.}
    
    \small
    \setlength{\tabcolsep}{3pt} 
    \begin{tabularx}{\linewidth}{l *{6}{>{\centering\arraybackslash}X}}
        \toprule
        \textbf{Model} & \textbf{MOTA}$\uparrow$ & \textbf{IDF1}$\uparrow$ & \textbf{HOTA}$\uparrow$ & \textbf{FP}$\downarrow$ & \textbf{FN}$\downarrow$ & \textbf{IDs}$\downarrow$ \\
        \midrule
        \textbf{Original} & 61.8 & 67.3 & 54.6 & 103440 & \textbf{88901} & 5243 \\
        \textbf{Ours} & 68.2 & \textbf{71.5} & \textbf{56.5} & 45822 & 115389 & 3513 \\
        \textbf{Ours-50\%} & \textbf{68.5} & 71.4 & 55.9 & 22000 & 137776 & \textbf{3283} \\
        \textbf{Ours-70\%} & 65.9 & 68.3 & 53.3 & \textbf{18777} & 154453 & 3449 \\
        \bottomrule
    \end{tabularx}
    \label{table:original_ours_baseline}
    
    \vspace{1ex} 
    \parbox{\linewidth}{\small All results are obtained from the MOT challenge server under the private detector protocol. The best results are shown in \textbf{bold}.}
\end{table}

\section{Conclusion}
In this paper, we introduced a reconstruction-based pruning method designed for the acceleration and compression of complex MOT models. Our method demonstrates promising results in effectively compressing intricate model architectures.

Future work will focus on enhancing our pruning technique and applying it to other complex architectures.

\bibliographystyle{IEEEtran}
\bibliography{refs}

\begin{thebibliography}{10}
\providecommand{\url}[1]{#1}
\csname url@samestyle\endcsname
\providecommand{\newblock}{\relax}
\providecommand{\bibinfo}[2]{#2}
\providecommand{\BIBentrySTDinterwordspacing}{\spaceskip=0pt\relax}
\providecommand{\BIBentryALTinterwordstretchfactor}{4}
\providecommand{\BIBentryALTinterwordspacing}{\spaceskip=\fontdimen2\font plus
\BIBentryALTinterwordstretchfactor\fontdimen3\font minus \fontdimen4\font\relax}
\providecommand{\BIBforeignlanguage}[2]{{%
\expandafter\ifx\csname l@#1\endcsname\relax
\typeout{** WARNING: IEEEtran.bst: No hyphenation pattern has been}%
\typeout{** loaded for the language `#1'. Using the pattern for}%
\typeout{** the default language instead.}%
\else
\language=\csname l@#1\endcsname
\fi
#2}}
\providecommand{\BIBdecl}{\relax}
\BIBdecl

\bibitem{yrjanainen2020privacy}
J.~Yrj{\"a}n{\"a}inen, X.~Ni, B.~Adhikari, and H.~Huttunen, ``Privacy-aware edge computing system for people tracking,'' in \emph{2020 IEEE International Conference on Image Processing (ICIP)}.\hskip 1em plus 0.5em minus 0.4em\relax IEEE, 2020, pp. 2096--2100.

\bibitem{sun2020collaborative}
H.~Sun, Y.~Chen, A.~Aved, and E.~Blasch, ``Collaborative multi-object tracking as an edge service using transfer learning,'' in \emph{2020 IEEE 22nd International Conference on High Performance Computing and Communications; IEEE 18th International Conference on Smart City; IEEE 6th International Conference on Data Science and Systems (HPCC/SmartCity/DSS)}.\hskip 1em plus 0.5em minus 0.4em\relax IEEE, 2020, pp. 1112--1119.

\bibitem{park2023multi}
J.~Park, J.~Hong, W.~Shim, and D.-J. Jung, ``Multi-object tracking on {SWIR} images for city surveillance in an edge-computing environment,'' \emph{Sensors}, vol.~23, no.~14, p. 6373, 2023.

\bibitem{agrawal2024systematic}
H.~Agrawal, A.~Halder, and P.~Chattopadhyay, ``A systematic survey on recent deep learning-based approaches to multi-object tracking,'' \emph{Multimedia Tools and Applications}, vol.~83, no.~12, pp. 36\,203--36\,259, 2024.

\bibitem{he2023fastreid}
L.~He, X.~Liao, W.~Liu, X.~Liu, P.~Cheng, and T.~Mei, ``{FastReID}: A {PyTorch} toolbox for general instance re-identification,'' in \emph{Proceedings of the 31st ACM International Conference on Multimedia}, 2023, pp. 9664--9667.

\bibitem{zhiqiang2017review}
W.~Zhiqiang and L.~Jun, ``A review of object detection based on convolutional neural network,'' in \emph{2017 36th Chinese control conference (CCC)}.\hskip 1em plus 0.5em minus 0.4em\relax IEEE, 2017, pp. 11\,104--11\,109.

\bibitem{vestias2019survey}
M.~P. V{\'e}stias, ``A survey of convolutional neural networks on edge with reconfigurable computing,'' \emph{Algorithms}, vol.~12, no.~8, p. 154, 2019.

\bibitem{liu2019edge}
S.~Liu, L.~Liu, J.~Tang, B.~Yu, Y.~Wang, and W.~Shi, ``Edge computing for autonomous driving: Opportunities and challenges,'' \emph{Proceedings of the IEEE}, vol. 107, no.~8, pp. 1697--1716, 2019.

\bibitem{wang2019towards}
Z.~Wang, L.~Zheng, Y.~Liu, and S.~Wang, ``Towards real-time multi-object tracking,'' \emph{The European Conference on Computer Vision (ECCV)}, 2020.

\bibitem{meimetis2023real}
D.~Meimetis, I.~Daramouskas, I.~Perikos, and I.~Hatzilygeroudis, ``Real-time multiple object tracking using deep learning methods,'' \emph{Neural Computing and Applications}, vol.~35, no.~1, pp. 89--118, 2023.

\bibitem{zhang2021fairmot}
Y.~Zhang, C.~Wang, X.~Wang, W.~Zeng, and W.~Liu, ``{FairMOT}: On the fairness of detection and re-identification in multiple object tracking,'' \emph{International journal of computer vision}, vol. 129, pp. 3069--3087, 2021.

\bibitem{he2023structured}
Y.~He and L.~Xiao, ``Structured pruning for deep convolutional neural networks: A survey,'' \emph{IEEE transactions on pattern analysis and machine intelligence}, 2023.

\bibitem{cheng2023survey}
H.~Cheng, M.~Zhang, and J.~Q. Shi, ``A survey on deep neural network pruning-taxonomy, comparison, analysis, and recommendations,'' \emph{arXiv preprint arXiv:2308.06767}, 2023.

\bibitem{fang2023depgraph}
G.~Fang, X.~Ma, M.~Song, M.~B. Mi, and X.~Wang, ``{DepGraph}: Towards any structural pruning,'' in \emph{Proceedings of the IEEE/CVF conference on computer vision and pattern recognition}, 2023, pp. 16\,091--16\,101.

\bibitem{zhou2019objects}
X.~Zhou, D.~Wang, and P.~Kr{\"a}henb{\"u}hl, ``Objects as points,'' \emph{arXiv preprint arXiv:1904.07850}, 2019.

\bibitem{yu2018deep}
F.~Yu, D.~Wang, E.~Shelhamer, and T.~Darrell, ``Deep layer aggregation,'' in \emph{Proceedings of the IEEE conference on computer vision and pattern recognition}, 2018, pp. 2403--2412.

\bibitem{luo2017thinet}
J.-H. Luo, J.~Wu, and W.~Lin, ``{ThiNet}: A filter level pruning method for deep neural network compression,'' in \emph{Proceedings of the IEEE international conference on computer vision}, 2017, pp. 5058--5066.

\bibitem{welch1995introduction}
G.~Welch, G.~Bishop \emph{et~al.}, ``An introduction to the {Kalman} filter,'' 1995.

\bibitem{tsai2022mobilenet}
C.-Y. Tsai and Y.-K. Su, ``{MobileNet-JDE}: a lightweight multi-object tracking model for embedded systems,'' \emph{Multimedia Tools and Applications}, vol.~81, no.~7, pp. 9915--9937, 2022.

\bibitem{wu2021sort}
H.~Wu, C.~Du, Z.~Ji, M.~Gao, and Z.~He, ``{SORT-YM}: An algorithm of multi-object tracking with {YOLOv4-tiny} and motion prediction,'' \emph{Electronics}, vol.~10, no.~18, p. 2319, 2021.

\bibitem{mao2019efficient}
Y.~Mao, Z.~He, Z.~Ma, X.~Tang, and Z.~Wang, ``Efficient convolution neural networks for object tracking using separable convolution and filter pruning,'' \emph{IEEE Access}, vol.~7, pp. 106\,466--106\,474, 2019.

\bibitem{jung2020real}
I.~Jung, K.~You, H.~Noh, M.~Cho, and B.~Han, ``Real-time object tracking via meta-learning: Efficient model adaptation and one-shot channel pruning,'' in \emph{Proceedings of the AAAI Conference on Artificial Intelligence}, vol.~34, no.~07, 2020, pp. 11\,205--11\,212.

\bibitem{liu2024fasttrack}
C.~Liu, H.~Li, and Z.~Wang, ``{FastTrack}: A highly efficient and generic {GPU}-based multi-object tracking method with parallel {Kalman} filter,'' \emph{International Journal of Computer Vision}, vol. 132, no.~5, pp. 1463--1483, 2024.

\bibitem{chollet2017xception}
F.~Chollet, ``Xception: Deep learning with depthwise separable convolutions,'' in \emph{Proceedings of the IEEE conference on computer vision and pattern recognition}, 2017, pp. 1251--1258.

\bibitem{dendorfer2020mot20}
P.~Dendorfer, H.~Rezatofighi, A.~Milan, J.~Shi, D.~Cremers, I.~Reid, S.~Roth, K.~Schindler, and L.~Leal-Taix{\'e}, ``{MOT20}: A benchmark for multi object tracking in crowded scenes,'' \emph{arXiv preprint arXiv:2003.09003}, 2020.

\bibitem{shao2018crowdhuman}
S.~Shao, Z.~Zhao, B.~Li, T.~Xiao, G.~Yu, X.~Zhang, and J.~Sun, ``{CrowdHuman}: A benchmark for detecting human in a crowd,'' \emph{arXiv preprint arXiv:1805.00123}, 2018.

\bibitem{milan2016mot16}
A.~Milan, L.~Leal-Taix{\'e}, I.~Reid, S.~Roth, and K.~Schindler, ``{MOT16}: A benchmark for multi-object tracking,'' \emph{arXiv preprint arXiv:1603.00831}, 2016.

\bibitem{bernardin2008evaluating}
K.~Bernardin and R.~Stiefelhagen, ``Evaluating multiple object tracking performance: the {CLEAR MOT} metrics,'' \emph{EURASIP Journal on Image and Video Processing}, vol. 2008, pp. 1--10, 2008.

\bibitem{ristani2016performance}
E.~Ristani, F.~Solera, R.~Zou, R.~Cucchiara, and C.~Tomasi, ``Performance measures and a data set for multi-target, multi-camera tracking,'' in \emph{European conference on computer vision}.\hskip 1em plus 0.5em minus 0.4em\relax Springer, 2016, pp. 17--35.

\bibitem{luiten2021hota}
J.~Luiten, A.~Osep, P.~Dendorfer, P.~Torr, A.~Geiger, L.~Leal-Taix{\'e}, and B.~Leibe, ``{HOTA}: A higher order metric for evaluating multi-object tracking,'' \emph{International journal of computer vision}, vol. 129, pp. 548--578, 2021.

\end{thebibliography}

\end{document}